\documentclass[10pt,twocolumn,letterpaper]{article}

\usepackage{cvpr}
\usepackage{times}
\usepackage{epsfig}
\usepackage{graphicx}
\usepackage{amsmath}
\usepackage{amssymb}
\usepackage{subfigure}
\usepackage[symbol]{footmisc}

\usepackage[breaklinks=true,bookmarks=false]{hyperref}

\cvprfinalcopy 

\newcommand{\norm}[1]{\|#1\|}

\begin{document}

\title{Reducing Racial Bias in Facial Age Prediction using Unsupervised Domain Adaptation in Regression}

\author{Apoorva Gokhale\thanks{Equal contribution. Names are listed alphabetically.} \qquad Astuti Sharma\footnotemark[1] \qquad Kaustav Datta\footnotemark[1] \qquad Savyasachi\footnotemark[1]\\
University of California, San Diego\\
{\tt\small \{agokhale, asharma, kdatta, ssavyasa\}@eng.ucsd.edu}

}

\maketitle
\begin{abstract}
We propose an approach for unsupervised domain adaptation for the task of estimating someone's age from a given face image. In order to avoid the propagation of racial bias in most publicly available face images datasets into inefficacy of models trained on them, we perform domain adaptation to motivate the predictor to learn features that are invariant to ethnicity, enhancing the generalization performance across faces of people from different ethnic backgrounds. Exploiting the ordinality of age, we also impose ranking constraints on the prediction of the model and design our model such that it takes as input a pair of images, and outputs both the relative age difference and the rank of the first identity with respect to the other in terms of their ages. Furthermore, we implement Multi-Dimensional Scaling to retrieve absolute ages from the predicted age differences from as few as two labelled images from the domain to be adapted to. We experiment with a publicly available dataset with age labels, dividing it into subsets based on the ethnicity labels, and evaluating the performance of our approach on the data from an ethnicity different from the one that the model is trained on. Additionally we impose a constraint to preserve the sanity of the predictions with respect to relative and absolute ages, and another to ensure the smoothness of the predictions with respect to the input. We experiment extensively and compare various domain adaptation approaches for the task of regression.
\end{abstract}

\section{Introduction}

Human beings are a wonderful species comprising of highly diverse races, each beautiful in their own way. Everyone has distinct facial features and it is very difficult, even for humans, to just judge someone's age by looking at their face. Imagine how difficult it would be for machines. There are numerous tasks where age prediction models could be useful. There are certain scenarios where we have age restrictions due to the services being offered not being suitable to certain age groups. For instance, certain content - both online and in the physical world is restricted for children and it is useful to have some automated tools which could could facilitate such access restrictions. We choose this problem due to its challenging nature and moral \& ethical applications.

There are prior works\cite{MLAR} which deal with this problem. However, just like any other machine learning task, the accuracy of the predictions of these models is limited by the datasets they are trained on. It is a well known fact that most face images datasets are biased towards Caucasian ethnicity\cite{CAUCBIAS} due to the ease of availability of the data and other races like Asians, American-African, South-East Asians are under represented in these datasets. As a result, the models trained on these datasets show poorer performance on the under-represented ethnicities. Furthermore, labeling an age dataset is cumbersome and expensive as compared to something like labeling genders. Thus, it is difficult to create sufficiently large labeled datasets for different. In our work, we try to tackle these problems and reduce the racial bias in age prediction models using the rapidly evolving techniques of domain adaptation. To the best of our knowledge, no one has tried to apply domain adaptation for the task of age prediction and this further strengthens the relevance of our work. 

Domain adaptation techniques help in reducing the domain bias introduced in the model trained on biased datasets. In a standard domain adaptation task, there is a single source domain for which ample labeled data is available and one or multiple target domains where the labeled data is very limited(Semi-Supervised Domain Adaptation) or not available at all (Unsupervised Domain Adaptation) and the objective is to leverage the available data to build a domain invariant predictor. We discuss this in detail in the next section.  

We model this problem as an unsupervised domain adaptation task where we try to adapt a deep convolutional age regressor trained on labeled data where identities are of Caucasian ethnicity to data from other ethnicities. We employ both Adversarial and Maximum Mean Discrepancy-based methods for domain adaptation. 

Since different races age differently, we think that the domain shift in an age prediction model might be too large to adapt. In order to tackle this, we try a different approach where we train a model to estimate the difference in the age, and also the ordering, on two given face images and then adapt it across ethnicities. We use Multi- Dimensional Scaling to convert the age differences predicted by the model to actual age values. This is driven by the notion that the features learned by the model which capture the difference in the age and the ordering should be more adaptable across different races as compared to the features which capture the absolute age. 

In this paper, first we discuss the Related Work in Section 2. In Section 3, We discuss various methods that we employ in detail. Section 4 comprises of the implementation details and different experiments we run in terms of hyper-parameters and variations in the proposed methods. We conclude and discuss future works and improvements in Section 5.   

\section{Related Work}
\subsection{Age prediction}
A number of existing works tackle the problem of age estimation from images. 
A classical machine learning approach, \cite{MLAR} perform coarse to fine prediction of age from images by extracting features such as Bio-Inspired Features (BIF), Kernel-based Local Binary Patterns (KLBP) and Multi-scale Wrinkle Patterns (MWP) that are used for Support Vector Regression.
\vspace{-10pt}
\paragraph{}In \cite{DEX} the authors use a VGG-16 \cite{VGG} backbone along with an expectation computed over the softmax output over all possible ages.
\vspace{-10pt}
\paragraph{}Inspired by \cite{DEX}, the authors of \cite{SSR} propose a novel network structure, and a compact stage-wise model for age estimation in which a dynamic range to each age group is introduced. The age interval of each group can be shifted and scaled depending on the input face image in this method.
\vspace{-10pt}
\paragraph{}The authors of \cite{DAREG} propose an approach for domain adaptation for age estimation in which they introduce a Gaussian kernel MMD-based and Graph Laplacian-based term in the objective function, to ensure that the model learns features that are invariant and are such that the predictions of the model preserve the smoothness in the inputs, across domains.

\subsection{Combined Regression and Ranking}
The author of \cite{CRR} proposes the joint optimization of both ranking and regression objective functions for enhancing the accuracy of the prediction ensuring good performance in both the tasks. This is implemented using a combined loss function that weighs squared error between the predicted and ground truth values and a logistic loss for the predicted rank and the actual ordering of a pair of values sampled from the same dataset. It is observed in the experiments conducted by him that optimizing upon the ranking objective along with the regression objective helps improve the model performance on predicting target values on rarely-occurring examples. In order to exploit this in our problem scenario we design our model to output both the difference between the ages of the identities in the two input images, and the rank which is a value from {0,1} depending on which identity is older. We optimize on a combined objective function taking both of these into account.

\subsection{Learning Distance Metrics between Pairs of Images}

In \cite{DML}, the authors tackle the problem of unsupervised domain adaptation when the source and the target domains have disjoint label spaces by formulating the classification problem into a verification task. They propose a \textit{Feature Transfer Network}, allowing simultaneous optimization of domain adversarial loss and domain separation loss, as well as a variant of N -pair metric loss for entropy minimization on the target domain where the ground-truth label structure is unknown, to further improve the adaptation quality. They demonstrate this for cross-ethnicity face verification that overcomes label biases in training data. Our approach is loosely related to this as we too try to estimate the difference between an ordinal attribute of two images input to the model simultaneously.

The conditional adversarial autoencoder (CAAE) proposed in \cite{UTK} achieves face age progression and regression in a holistic framework. Starting from an arbitrary query face without knowing its true age, they are able to freely produce faces at different ages, while at the same time preserving the personality. 

DEX, in \cite{DEX}, tackled the estimation of apparent age in still face images.It posed the age regression problem as a deep classification problem followed by a softmax expected value refinement and show improvements over direct regression training of CNNs. DEX ensembles the prediction of 20 networks on the cropped face image. DEX does not explicitly employ facial landmarks. The paper also crawled internet face images with available age to create a large public dataset, IMDB-WIKI. However the dataset doesn't have ethnicity labels making the dataset not directly usable for our objective.


\section{Methods}
We focus on unsupervised domain adaptation. We have a source dataset $X_s = \{x_i , y_i\}_{i=0}^{N_s}$ drawn from a labeled source domain S, and a dataset $X_t = \{x_j\}^{Nt}$ from a different unlabeled target domain. For our case, source and target domains are images from different ethnicities and labels are age values. As we treat age as continuous values, we model this as a regression task. All of our models are deep neural networks with a regression layer plugged in the end. 
\\
We experiment with some baseline models, and further built our approach on them. In next few sections we talk about our models and motivation of using them along with details of their architecture.
\\
\subsection{Baseline: Source Only}
For a baseline of our work, we work with only single domain age predictor models. We take a pre-trained model, and remove the classifier to plug in our regression module and then fine-tune it to predict age values. The regression module consists of multiple linear layers with a single output in the end for giving out age values. We experiment with MSE(L2) and MAE(L1) loss and find better performance with MAE loss and therefore, use MAE loss for all our regression models. We try different pre-trained models such as different variants of ResNet\cite{RESNET} and Inception\cite{INCEPTION} and in our case, ResNet-50 gives the best performance and we use that as our base model for all of our experiments.  
\\

\subsection{Pairwise image ranking} 
We adopt an alternative approach, where our model learns to predict the relative difference in ages between a pair of images, and their ranking. The pair of images are concatenated and fed into the network as a 6 channel input. The base architecture remains the same as our baseline model above, except now instead of the single output at the end of our regression module, we have two outputs representing the distance between the images, and our ranking value that comes from a sigmoid activation. The loss function consists of two terms, namely the L1/L2 regression loss, and a binary cross entropy loss for ranking. As mentioned previously, this ranking information aids our model to adapt better because it is reasonable to assume that age differences remain consistent across ethnicties. \\
Now that we can compute the pairwise distance between any two images, to obtain our final predictions of the absolute ages, we employ Multi Dimensional Scaling, which is explained in the next section.

\subsection{Multi Dimensional Scaling}
Multi Dimensional Scaling (MDS) is an algorithm that, given pairwise dissimilarities between each pair of objects in a set and a specific number of dimensions N, maps the original objects into this N dimensional space such that the pairwise distances are preserved as best as possible. We use Metric MDS, which works by minimizing a cost function called Stress, defined as below -
\begin{equation}
    Stress_D(x_1, x_2, ... x_n) = \sqrt{\sum_{i \neq j =1,..N}(d_{ij} - \norm{x_i - x_j})^2}
\end{equation}
where $x_1,...x_n$ are the mapped points in N dimensional space, and $d_{ij}$ is an element of the distance matrix.\\

\subsection{Label Normalization}
In order to make the gradients smoother and more stable while training the model, we experimented with label normalization to limit the output labels, absolute ages or age differences, to the range [0,1].
Empirically, we found that this does not have a positive influence on the training process.

\subsection{Identity Constraints}
It is desirable that the predictor $f$ (or $f_1$ or $f_2$) is a reasonable predictor, we choose to exploit the regularity the age difference between images that are copies of each other should be 0 and the ranking function should have maximum uncertainty, and the age difference output by the model $f_1(A,B) = - f_1(B,A)$, $f_1$ being the predictive function for age difference. Also, $f_2(A,B) = f_2(B,A)'$ , $f_2$ being the ranking function. We include this in our complete objective function.

\subsection{Adaptive Measures}
\subsubsection{Adversarial}
 One approach to domain adaptation is to learn the domain invariant representation of the data. That is, we learn a model that can generalize well from one domain to another, and thus finding a internal representation which contains no discriminative information about the origin of the input (source or target), while also preserving a low risk on the source (labeled) examples. Based on this idea we use DANN as one of our domain adaptation methods.\\
 DANN(Domain Adversarial training of Neural Networks) implements the idea discussed above in the context of neural network architectures that are trained on labeled data from the source domain and unlabeled data from the target domain. As the training progresses, the approach promotes the emergence of features that are (i) discriminative for the main learning task on the source domain and (ii) indiscriminate with respect to the shift between the domains. This adaptation behaviour can be achieved in almost any feed-forward model by augmenting it with few standard layers and a new gradient reversal layer. The resulting augmented architecture can be trained using standard backpropagation and stochastic gradient descent. The complete DANN architecture is shown in Figure1.
 \\
 We use ResNet as the feature extactor and our regression module as label predictor which together form a standard feed-forward architecture. Unsupervised domain adaptation is achieved by adding an adverserial network which is composed of three linear layers, connected to the feature extractor via a gradient reversal layer that multiplies the gradient by a certain negative constant during the backpropagation-based training.
 \\
\begin{figure}[!htbp]
\begin{center}
\includegraphics[scale=.3]{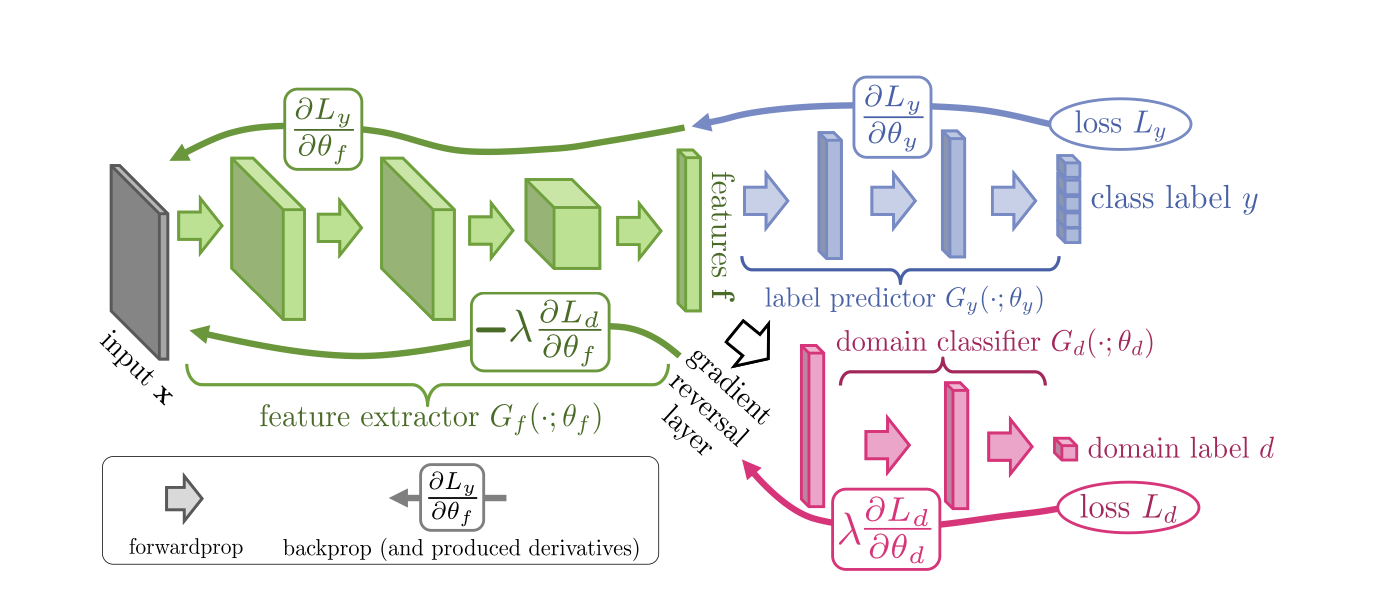} 
\end{center}
\caption{Domain Adverserial training of Neural Networks (DANN)}
\end{figure}

\subsubsection{MMD}
Due to the unavailability of labels in the target domain, one commonly used strategy of UDA is to learn domain invariant representation via minimizing the domain distribution discrepancy. On this idea, many existing methods aim to bound the target error by the source error plus a discrepancy metric between the source and the target. One such discrepancy metric MMD(Maximum Mean Discrepancy) is based on the notion of embedding probabilities in a reproducing kernel Hilbert space. It measures distance between kernel mean embeddings of two distributions. For mmd, we use gausian kernel. We do single layer adaptation on fc1. 

\subsubsection{Smoothing}
In unsupervised domain adaptation for classification tasks, the unlabeled target data is often utilized to find the parameter configuration that leads to a classifier in the target domain that passes through low density regions in the input feature space. This is achieved by implementing entropy minimization, maximizing the classifier's confidence on the target domain inputs. Although we do not have a corresponding approach minimizing the uncertainty of regression outputs, we leverage Graph Laplacian-based smoothing, which tries to enforce that if inputs $x_i$ \& $x_j$ are close to each other in the input space, then the corresponding outputs $f(x_i)$ \& $f(x_j)$ should also be similar to each other. This smoothness is computed as:
\begin{center}
\begin{equation}
    f^TLf = \frac{1}{2} \sum_{i,j}w_{ij}(f(x_i)-f(x_j))^2
\end{equation}
\end{center}
where,
\begin{center}
\begin{equation}
    w_{ij} = exp(\frac{-||x_i - x_j||^2}{2\sigma^2})
\end{equation}
\end{center}

\subsubsection{Final Objective Function} 
The final objective function that we use to optimize our model is a combination of the L1 loss on the predictions on the labelled source domain data, the objective term associated with the domain adaptation measure (MMD or Adversarial). In case of the input being pairwise the L1 Loss is on the relative age difference predicted on the labelled data. We also include the Ranking Loss on the ranks output by the model for the source domain examples, the identity and inversion constraint loss and the smoothing loss. It can be represented as follows:
\vspace{-24pt}
\begin{center}
    \begin{equation}
        L = L_{reg}(f_1(g(y_s)),y_s) + L_{adv}(g(x_s),g(x_t))
    \end{equation}
\end{center}
or in case of pairwise inputs :
\vspace{-25pt}
\begin{center}
    \begin{multline}
        L = L_{reg}(f_1(g(x_{s,1}, x_{s,2})),y_{s,1} - y_{s,2}) \vspace{-6pt}\\ + \alpha L_{rank}(f_2(g(x_{s,1}, x_{s,2})),r(y_{s,1} - y_{s,2}))\vspace{5pt} \\ + \gamma L_{adv}(g(x_{s,1},x_{s,2}),g(x_{t,1},x_{t,2})) \vspace{5pt}\\ + \beta L_{id}(f_1(g(x_{s,1}, x_{s,2})),f_1(g(x_{t,1}, x_{t,2}))) \vspace{5pt} \\ + \sigma L_{smooth}(g(x_s), g(x_t), f_1(g(x_s)), f_1(g(x_t)))
    \end{multline}
    \end{center}
where $r$ is the ranking function, $g$ is the feature extractor and $f_1$ is the regression output function and $f_2$ is the ranking output function.

\begin{table}[!htbp]
\caption{Loss(MAE) for various domain combination using baseline model}
\begin{center}
\begin{tabular}{|l|l|l|l|}
\hline
{\bf Source}&{\bf Target}&{\bf Source Val Loss}&{\bf Target Loss}\\
\hline
Caucasian & African & 7.232 & 7.946 \\ 
\hline
Caucasian & Asian & 8.122 & 7.677 \\
\hline
\end{tabular}
\end{center}
\end{table}

\section{Experiments}

\subsection{Dataset} 
In this work, we run experiments on the UTK\cite{UTK} dataset. UTKFace dataset is a large-scale face dataset with long age span (range from 0 to 116 years old). The dataset consists of over 20,000 face images with annotations of age, gender, and ethnicity. The images cover large variation in pose, facial expression, illumination, occlusion, resolution, ethnicity etc. We use the cropped and aligned face image variant of the dataset. The  Figure [] shows some of the examples of processed face images from the dataset of different ethnicities. The images are resized to size 200x200 and we also employ random horizontal flips for data augmentation.However, as seen in Table \ref{t:datadist}, there is a large bias even in this dataset in terms of the distribution of the images among the 5 ethnicities.

\begin{table}[!htbp]
\caption{Distribution of Images among Ethnicities}
\vspace{10pt}
\label{t:datadist}
\begin{center}
\begin{tabular}{|l|l|}
\hline
{\bf Ethnicity} &{\bf Number of Images} \\
\hline
 Caucasian & 10,078 \\
 \hline
 African & 4,526\\
\hline
 Asian & 3,434 \\
\hline
 Indian & 3,975 \\
\hline
 Others & 1,692 \\
\hline
\end{tabular}
\end{center}
\end{table}

\begin{figure}
\hspace{0.1\textwidth}
\begin{subfigure}
  \centering
  \includegraphics[width=.2\linewidth]{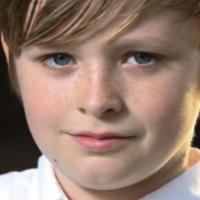}
\end{subfigure}%
\hspace{0.\textwidth}
\begin{subfigure}
  \centering
  \includegraphics[width=.2\linewidth]{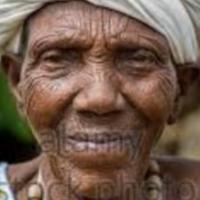}
\end{subfigure}\\

\hspace{0.1\textwidth}
\begin{subfigure}
  \centering
  \includegraphics[width=.2\linewidth]{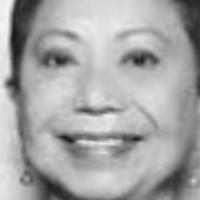}
\end{subfigure}
\hspace{-2pt}
\begin{subfigure}
  \centering
  \includegraphics[width=.2\linewidth]{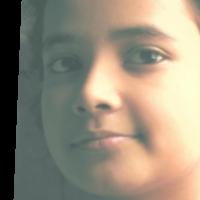}
\end{subfigure}\\

\hspace{0.155\textwidth}
\begin{subfigure}
  \centering
  \includegraphics[width=.2\linewidth]{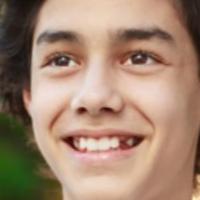}
\end{subfigure}%
\caption{Example images of different ethnicities from the UTK dataset}
\label{fig:dataset_imgs}
\end{figure}

\subsection{Implementation Details}
All of the code is written using PyTorch. We find that learning rate 1e-3 gives the best convergence and we use that for all our models. All of our models are trained using stochastic gradient descent on a single GPU with a batch size of 16. We use Adam \cite{ADAM} for our regression network and SGD for the adversarial network. 


\subsection{Pairwise Models}
This section describes our experiments where inputs to the model are pairs of images of faces of the same ethnicity.

\subsubsection{Rank and Without Rank}
We experiment to find the usefulness of rank in driving the regression outputs of the model to better values. We find that augmenting the original L1 Loss on the age difference and its prediction with a weighed binary cross entropy loss on the predicted rank indeed improves the performance of the model on the regression task. While the original model trained only on the L1 Loss achieves minimum L1 Loss of 10.15 on the training set and 12.67 on the validation set, the respective L1 Loss values for the ranking-augmented regression approach are 6.23 and 7.9. This shows that adding the ranking objective enhances the accuracy of the regression outputs from the classifier sharing weights up until the bottleneck layer.

\subsection{Adaptive Models}

\begin{table}[!htbp]
\caption{Loss(MAE) with different layers adaptation for baseline model for Caucasian to African}
\label{t:layers}
\begin{center}
\begin{tabular}{|l|l|l|}
\hline
{\bf Layers} &{\bf S Val Loss}&{\bf Target Loss} \\
\hline
 No Adaptation & \textbf{6.326} & 6.986 \\
 \hline
 conv+fc123 & 6.384 & \textbf{6.876}\\
\hline
 conv+fc1 & 7.056 & 8.150 \\
\hline
 fc1&  7.144 & 7.728 \\
\hline
 fc123& 7.356& 7.876 \\
\hline
\end{tabular}
\end{center}
\end{table}

\subsubsection{Layers}
We experimented with single and multiple layers adaptation for both of our adaptative measures, DANN and MMD. We take into account both last convolutional layer of our feature extractor and fully connected layers from our regression module. With experiments we find that adapting convolutional layer and first fully connected layer gave better results as compared to single or different combinations of multiple layers. This experiment is performed on the standard age prediction model(not pairwise). Table \ref{t:layers} summarizes the results.

\begin{table}[!htbp]
\caption{Loss(MAE) with different DANN hyperparameter $\gamma$ on pre-trained model for Caucasian to African}
\begin{center}
\begin{tabular}{|l|l|l|}
\hline
{\bf $\gamma$ value} &{\bf S Val Loss}&{\bf Target Loss} \\
\hline
 0.1 &  \textbf{6.582} & \textbf{7.787} \\
\hline
 1 &  7.281 & 8.256 \\
\hline
\end{tabular}
\end{center}
\label{t:pre-trained}
\end{table}
\subsubsection{Pre-Trained Regression model}
Since, The adaptive models were giving poor prediction results, In this experiment we first train a baseline (source only) model till it starts showing good performance on the source data. We then take this pre-trained model and plug in adaptation using an adversarial network. We try $0.1$ and $1$ values for $\gamma$ - the hyperparameter for adversarial loss and the table and figure summarizes the results for the same. For $\gamma = 1$, we find that as the model tries to make features domain invariant, it loses its regressive power and thus performs poorly on both source and target domain as compared to even the baseline model. Fir $\gamma = 0.1$, we conclude that signal is not strong enough to give any significant boost in the overall performance. Table \ref{t:pre-trained} summarizes the results.

\subsubsection{DANN vs MMD}
We experiment with two domain adaptation measures as discussed in Section 3.4 in detail. According to our obtained results, using adversarial loss performs better than MMD. We observe that the target loss for DANN is lower than that of MMD after training the network for adaptation.

\begin{table}[!htbp]
\caption{Loss(MAE) for DANN and MMD(fc1) on pre-trained model for Caucasian to African}
\begin{center}
\begin{tabular}{|l|l|l|}
\hline
{\bf Method} &{\bf S Val Loss}&{\bf Target Loss} \\
\hline
 DANN & 7.144 & 7.728 \\
\hline
 MMD &  8.136 & 8.393 \\
\hline
\end{tabular}
\end{center}
\label{t:pre-trained}
\end{table}

\begin{table}[!htbp]
\caption{Loss(MAE) with different DANN hyperparameter $\gamma$ on baseline model for Caucasian to African}
\begin{center}
\begin{tabular}{|l|l|l|}
\hline
{\bf $\gamma$ value} &{\bf S Val Loss}&{\bf Target Loss} \\
\hline
 0.1 &  \textbf{6.812} & 7.755 \\
\hline
 0.3 & 7.086 &\textbf{7.591} \\
\hline
 0.6 & 6.896 &7.749 \\
\hline
 1 &  7.232 & 7.946 \\
\hline
\end{tabular}
\end{center}
\label{t:gamma}
\end{table}

\subsubsection{Adversarial Hyper-Parameter}
In order to control the effect of adversarial domain adaptation loss, we try different values of the hyperparameter $\gamma$ and the Table \ref{t:gamma} summarizes the results. We adapt all the layers in this scenario and use the standard age prediction model for adaptation. Similar to the results for the pre-trained model, we find that for larger values of $\gamma$, the gradient to make features domain invariant, the prediction capabilities of the model get compromised. However, for smaller values, It is too weak to show any improvement on the target set.   

\subsubsection{MDS}
We use two labelled images from our target domain as references, to find out the absolute values of the ages of our original images, after we map our pairwise distances to 1 component using Metric MDS. The two reference images are used for inferring relative difference of ages of two images after MDS, and using the information of their absolute values, we infer the absolute values of the other images. We report the loss values for MDS for two settings - with and without the ranking loss.

\begin{table}[!htbp]
\caption{Loss for MDS}
\vspace{10pt}
\label{t:datadist}
\begin{center}
\begin{tabular}{|l|l|}
\hline
{\bf Rank/Regression} &{\bf Target Loss}  \\
\hline
 Rank + Regression & 17.87 \\
 \hline
 Regression & 18.67\\
\hline
\end{tabular}
\end{center}
\end{table}

\section{Conclusion and Future Work}
We tackle the problem of age estimation based on face images and then try to reduce the racial bias in the predictor using domain adaptation techniques. We train a deep convolutional regression model for age estimation and then use domain adaptation techniques(adversarial and MMD) to improve the generalization performance. We model the age estimation in two ways - as a standard regression problem and as a pairwise approach where the model is trained to predict the age difference between a pair of face images and then we use Multi-Dimensional Scaling to find the absolute age values. We find that even though the features may be more adaptable in a model which captures the difference in ages among two face images, the prediction capability of the model is lesser in comparison to the standard regression model. We think that this may be due to the overall difficulty of the predicting the difference in age as compared to predicting the absolute age since now there could be more variation in the pair of faces and also an additional step of MDS. In terms of adaptation, Adversarial techniques tend to give better performance than MMD based ones. However the difference is not much. We also find that, as we try to adapt features from multiple layers, the features become more domain invariant but the regression capability gets compromised. Overall, we find that domain adaptation across races for age estimation is a challenging task. After adaptation, we see some improvements in the perforce, e.g. our baseline with conv+fc1 target MAE loss reduces by value of 0.11(as compared to source only performance). However in totality, by observing performance of all our experiments, we don't see any major improvement in the performance on the target domain. We observe that when the adaptation feedback is strong, the model loses its general prediction capabilities and we see a drop in performance in both source and target domains. Also, if the signal is too weak, there's not much improvement in terms of adaptation. Further hyper-tuning of the model may result in a boost in performance. The performance is also limited due to the dataset we're using currently which may not be large enough for the problem at hand. There are not many datasets available with ethnicity labels. In future works, we plan on creating an ethnicity labeled dataset from existing face images dataset but first training an ethnicity classifier on the given dataset and then using that to infer the ethnicity labels. Domain mapping can also be implemented here, using generative modelling to synthesize images of one ethnicity from another such that the attributes that are telling for age are preserved. Other directions for future improvement can be making use of multi-modal inputs such as sound along with images, so that the hoarseness and timbre of the voice can be used as a feature to estimate the age of the person.



\bibliographystyle{ieee}

\end{document}